\documentclass{article}
\pdfoutput=1
\usepackage{arxiv}

\usepackage[utf8]{inputenc} 
\usepackage[T1]{fontenc}    
\usepackage{url}            
\usepackage{booktabs}       
\usepackage{amsfonts}       
\usepackage{amsmath}

\usepackage{mathabx}

\usepackage{nicefrac}       
\usepackage{microtype}      
\usepackage{lipsum}
\usepackage{graphicx}
\usepackage[natbibapa]{apacite}
\graphicspath{ {./images/} }
\usepackage{bbm} 
\newcommand{\eqdef}{\mathrel{\mathop:}=}
\newcommand{\rs}{\mathbf{s}}
\newcommand{\ra}{\mathbf{a}}
\newcommand{\rst}{\mathbf{s}_{1:T}}
\newcommand{\rat}{\mathbf{a}_{1:T}}
\newcommand{\rsh}{\mathbf{s}_{t:T}}
\newcommand{\rah}{\mathbf{a}_{t:T}}
\newcommand{\optimal}{\mathcal{O}}

\newcommand{\KL}{D_\mathrm{KL}}
\newcommand{\E}{\mathbb{E}}

\usepackage{wrapfig}

\usepackage{xcolor}
\usepackage{graphicx}
\usepackage{subcaption}
\title{Reinforcement Learning as Iterative and Amortised Inference}

\author{
  Beren Millidge* \\
  School of Informatics\\
  University of Edinburgh \\
  \texttt{beren@millidge.name} 
   \And
   Alexander Tschantz* \\
   Sackler Centre for Consciousness Science \\
   School of Engineering and Informatics \\
   University of Sussex \\
   \texttt{tschantz.alec@gmail.com}
   \And
   Anil K Seth \\
   Sackler Centre for Consciousness Science \\
   Evolutionary and Adaptive Systems Research Group \\
   University of Sussex \\
   \texttt{A.K.Seth@sussex.ac.uk}
   \And
   Christopher L Buckley \\
   Evolutionary and Adaptive Systems Research Group\\
   School of Engineering and Informatics \\
  University of Sussex \\
   \texttt{C.L.Buckley@sussex.ac.uk}}

\begin{document}

\maketitle

\begin{abstract}
    There are several ways to categorise reinforcement learning (RL) algorithms, such as either model-based or model-free, policy-based or planning-based, on-policy or off-policy, and online or offline. 
    Broad classification schemes such as these help provide a unified perspective on disparate techniques and can contextualise and guide the development of new algorithms. 
    In this paper, we utilise the control as inference framework to outline a novel classification scheme based on amortised and iterative inference. 
    We demonstrate that a wide range of algorithms can be classified in this manner providing a fresh perspective  and highlighting a range of existing similarities.
    Moreover, we show that taking this perspective allows us to identify parts of the algorithmic design space which have been relatively unexplored, suggesting new routes to innovative RL algorithms \footnote{* Equal Contribution}.
\end{abstract}

\section{Introduction}
\label{sec:introduction}

Many classification schemes exist for reinforcement learning (RL) algorithms. Algorithms can be classified as either model-based or model-free, depending on whether a model of the environment is utilised. Alternatively, RL algorithms can be classified as either  policy-based or planning-based, on-policy or off-policy, and online or offline.
These classification schemes help provide a unified perspective on RL, highlighting similarities and differences amongst approaches and aiding the development of novel algorithms.

In this work, we highlight a relatively uncharted classification scheme based on \emph{iterative} and \emph{amortised}  inference.  
Inspired by the control as inference (CAI) framework \cite{dayan1997using, rawlik2010approximate, toussaint2006probabilistic, ziebart2010modeling, levine2018reinforcement, fellows2019virel}, we cast the problem of reward maximization in terms of variational inference. 
In this context, iterative inference approaches directly optimize the posterior distribution, while amortised methods learn a parameterised function (e.g., a policy, or amortised value function) which maps directly from states to the quantity of interest (such as actions or Q-values).
 
We demonstrate that this classification scheme provides a principled partioning of a wide range of existing approaches to RL, including policy gradient methods, Q-learning, actor-critic methods, trajectory optimisation and stochastic planning, and that by doing so it provides a novel perspective which highlights algorithmic commonalities that may  otherwise be overlooked. We find that existing implementations of iterative inference generally correspond to model-based planning, whereas implementations of amortised inference generally correspond to model-free policy optimisation. Importantly, this classification scheme highlights unexploited regions of algorithmic design space in \textit{iterative policies} and \textit{amortised plans}, and also in the combination of iterative and amortised methods. Exploring these new regions has the potential to inspire novel RL algorithms. 

\section{Control as Inference}
\label{sec:cai}

We consider a Markov Decision Process (MDP) defined by $\{ \mathcal{S}, \mathcal{A}, p_{\texttt{env}}, r, \gamma \}$, where $\ra \in \mathcal{A}$ denotes actions and $\rs \in \mathcal{S}$ denotes states. 
State transitions are governed by $\rs_{t+1} \sim p_{\texttt{env}}(\rs_{t+1}|\rs_{t}, \ra_{t})$, and the reward function is $r(\rs_t, \ra_t)$. $\gamma \in (0, 1]$ is a factor which discounts the sum of rewards $r(\tau) = \sum_t \gamma^t r(\rs_t, \ra_t)$, where $\tau$ denotes a trajectory $\tau = (\rs_1, \ra_1, ..., \rs_T, \ra_T)$. 
RL aims to optimise a policy distribution $p(\ra_t|\rs_t)$. 
The probability of trajectories under this policy is given by $p_{\pi}(\tau) = p(\rs_1) \prod_{t=1}^T p_{\texttt{env}}(\rs_{t+1}|\rs_t, \ra_t)p=(\ra_t|\rs_t)$.
In traditional RL, the objective is to maximise the expected sum of returns $\E_{p_{\theta}(\tau)}[r(\tau)]$ (throughout we assume $\gamma = 1$). 

To reformulate this objective in terms of probabilistic inference, we  construct a graphical model where the posterior distribution over actions $p(\ra_t|\rs_t)$ recovers the optimal policy. 
This requires the graphical model to incorporate some notion of reward, which is achieved by introducing an additional binary variable $\mathcal{O}$, where $\mathcal{O} \in [0, 1]$, which is referred to as an `optimality' variable, as $\mathcal{O}_t = 1$ implies time step $t$ was `optimal'. Since we only ever desire optimality, we will drop it from the notation: $p(\optimal_t =1) \eqdef p(\optimal_t)$.
The corresponding graphical model is shown in the Appendix.

The  objective of CAI is to obtain the optimal posterior $p(\ra_{t:T} | \rs_{t:T}, \optimal_{t:T})$ over a full trajectory. There are many ways to approximate this desired posterior, ranging from variational inference \citep{wainwright2008graphical,beal2003variational}, message passing algorithms \citep{yedidia2011message,weiss2000correctness}, and importance sampling \citep{kappen2016adaptive,kutschireiter2020hitchhiker}. Each of these approaches can be seen to correspond to a family of RL algorithms in the literature. There are also two approaches to optimising the posterior $p(\ra_{t:T} | \rs_{t:T}, \optimal_{t:T})$. One can either directly optimize it, in which case one infers a full sequence of actions $\ra_{t:T}$ -- i.e. a \emph{plan} -- or else one can choose to infer a sequence of single-step action posteriors $p(\ra_t | \rs_t, \optimal_{t:T})$ which corresponds to sequentially inferring \emph{policies}.

One approach to approximating the true posterior, which underlies a large subset of common RL algorithms, is through variational inference. Here, we introduce an approximate posterior $q(\tau;\theta) = q(\rs_1) \prod^T_{t=1} q(\rs_{t+1}|\rs_t, \ra_t)q(\ra_t|\rs_t;\theta)$, and use this to construct a variational bound $\mathcal{L}$ on the true-posterior. Let $p(\tau, \mathcal{O}_{1:T})$ denote an agent's generative model, which can be factorized as:
\begin{align*}
    p(\tau, \mathcal{O}_{1:T}) = p(\rs_1) \prod^T_{t=1} p(\mathcal{O}_t|\rs_t, \ra_t) p(\rs_{t+1}|\rs_t, \ra_t)p(\ra_t)
\end{align*}

The likelihood of optimality is usually defined as $p(\mathcal{O}_t = 1|\rs_t, \ra_t) = \mathrm{exp}(\frac{1}{\beta}r(\rs_t, \ra_t))$, thereby maintaining consistency with traditional RL objectives. Here, $\beta$ is a temperature parameter which scales the contribution of the reward and entropy terms. Traditional RL algorithms are recovered as $\beta \rightarrow 0$. We additionally assume an uninformative uniform action prior. Given these definitions, the variational bound $\mathcal{L}$ is defined as (see Appendix for a derivation and discussion of the assumptions required).
\begin{equation}
\begin{aligned}
\label{eq:objective}
    \mathcal{L} &\eqdef \KL\Big(q(\tau;\theta) \Vert p(\tau, \mathcal{O}_{1:T})\Big) \leq \KL\Big(q(\tau;\theta) \Vert p(\tau| \mathcal{O}_{1:T})\Big)\\
\end{aligned}
\end{equation}
\begin{equation}
\begin{aligned}
\label{eq:simple-objective}
\mathcal{L} = \E_{q(\tau;\theta)}\Big[\log p(\mathcal{O}_{1:T} = 1| \tau)\Big]  + \mathcal{H}\Big[q(\rat|\rst;\theta)\Big]
\end{aligned}
\end{equation}
where $\mathcal{H}[\cdot]$ is the Shannon entropy. 
Maximising $\mathcal{L}$ is thus equivalent to maximising both the expected likelihood of optimality and the entropy of $q(\rat|\rst; \theta)$. The inclusion of an entropy term over actions provides several benefits such as including a mechanism for offline learning \cite{nachum2017bridging,levine2020offline}, improving, and increasing algorithmic stability and robustness. Empirically, algorithms derived from the control as inference framework often outperform their non-stochastic counterparts \cite{haarnoja2018soft,hafner2018learning, hausman2018learning}.

\section{Iterative \& Amortised Inference}
\label{sec:iter-amort}
In the wider literature on probabilistic inference, a key distinction is made between \emph{iterative} and \emph{amortised} approaches to inference.
Iterative methods directly optimise the parameters of the approximate posterior, a process which is carried out for each data-point. While this inference procedure could theoretically be single-step, in practice most algorithms are iterative, hence the name. Examples of this method include belief propagation \citep{pearl2014probabilistic} variational message passing \cite{winn2005variational}, stochastic variational inference \cite{hoffman2013stochastic}, black box variational inference \citep{ranganath2013black}, and expectation-maximisation \citep{dempster1977maximum}.

In contrast, amortised approaches to inference \citep{kim2018semi,marino2018iterative,rezende2014stochastic}  learn a parameterised function $f_{\phi}(\rs_t)$ which maps directly from $\rs_t$ to the parameters of the approximate posterior $\theta \leftarrow f_{\phi}(\rs_t)$.
Amortised inference models are learned by optimising the parameters $\phi$ in order to maximise $\mathcal{L}$ over the available dataset $\mathcal{D}$. In practice, $f_{\phi}(\cdot)$ is often implemented as a neural network with weights $\phi$. Amortised inference forms the basis of variational autoencoders \cite{kingma2013auto}, one of the most popular tools for inference in machine learning. We use $q(\ra_t | \rs_t; \theta)$ to denote a variational posterior optimised through iterative inference with parameters $\theta$, and $q_{\phi}(\ra_t|\rs_t) = q(\ra_t; \theta = f(\rs_t; \phi))$ to denote an \emph{amortised} posterior with parameters (of the amortisation function) $\phi$. These two approaches optimise subtly different objectives, which we present below\footnote{For convenience, we showcase the distinction on the variational lower-bound derived earlier.}.

\begin{equation}
\begin{aligned}
\label{eq:iter-amort}
    &\textbf{\textit{\small{Iterative Inference Objective}}} \\ 
    &\mathrm{arg max}_{\theta} \E_{q(\tau;\theta)}\Big[\log p(\mathcal{O}_{1:T} = 1| \tau)\Big]  + \mathcal{H}\Big[q(\rat|\rst;\theta)\Big]\\
    &\textbf{\textit{\small{Amortised Inference Objective}}} \\
    &\mathrm{arg max}_{\phi} E_{\mathcal{D}}\Bigg[ \E_{q_\phi(\tau)}\Big[\log p(\mathcal{O}_{1:T} = 1| \tau)\Big] + \mathcal{H}\Big[q_{\phi}(\rat|\rst)\Big] \Bigg]
\end{aligned}
\end{equation}

\section{Classification Scheme}
\label{sec:class-scheme}
\label{ap:graphical-model}
\hspace{-5cm}
\begin{figure}
    \begin{center}
          \includegraphics[width=0.8\textwidth]{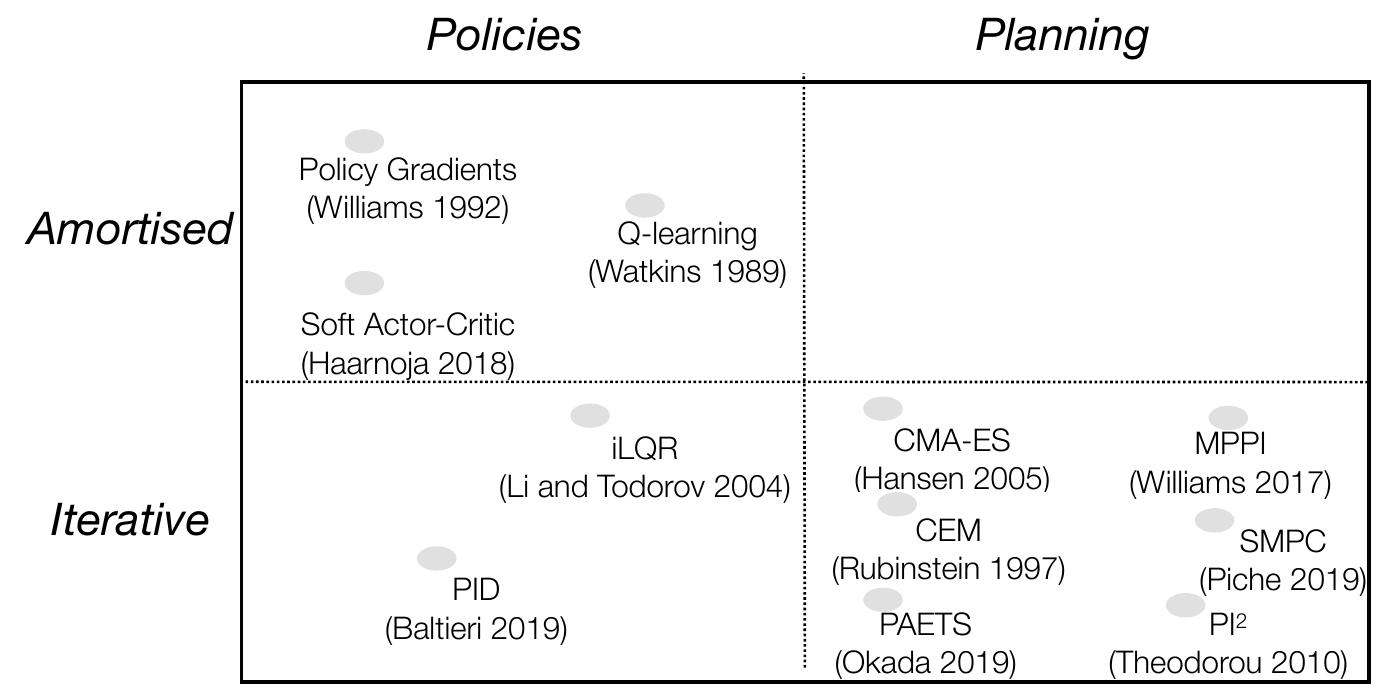}
    \end{center}
    \caption{Overview of classic RL and control algorithms in our scheme. Standard model-free RL corresponds to amortised policies, planning algorithms are iterative planning, and control theory infers iterative policies. The amortised plans quadrant is empty, perhaps suggesting room for novel algorithms.}
    \label{fig:quad}
\end{figure}

We propose a classification scheme which partitions RL algorithms along two orthogonal axes of variation -- whether they optimize \emph{plans} or \emph{policies}, and whether they utilize \emph{iterative} or \emph{amortised} inference. Below, we classify a number of established RL algorithms in terms of our scheme.

\paragraph{Policy Gradients} By directly differentiating the variational lower-bound, one can derive the policy-gradient class of algorithms \cite{sutton2000policy,schulman2017equivalence}. Typically, amortised methods are used and the optimisation is performed over the whole dataset. We can derive updates for $\phi$ by differentiating the amortised objective in Eq. \ref{eq:iter-amort} w.r.t $\phi$, $\nabla_{\phi}\E_{\mathcal{D}}\big[\mathcal{L}(\phi)\big]$:
\begin{equation}
\begin{aligned}
\label{eq:policy-gradients}
    &\nabla_{\phi}\E_{\mathcal{D}}\big[\mathcal{L}(\phi)\big] = \E_{\mathcal{D}}\big[\nabla_{\phi} \mathcal{L}(\phi)\big] = \E_{\mathcal{D}} \bigg[\nabla_{\phi} \E_{q(\tau)}\Big[ \sum_{t}^T r(\rs_t, \ra_t) - \log q_{\phi}(\ra_t|\rs_t) \Big]\bigg] \\
    &=\E_{\mathcal{D}} \bigg[\E_{q(\tau)}\Big[\nabla_{\phi} \log q_{\phi}(\ra_t|\rs_t) \Big(\sum_{t}^T r(\rs_t, \ra_t) - \log q_{\phi}(\ra_t|\rs_t) \Big) \Big]\bigg]
\end{aligned}
\end{equation}

Equation \ref{eq:policy-gradients} resembles the standard policy gradient objective, but with an additional entropy term over actions, which encourages exploration by rewarding entropic policies. 
\paragraph{Q-Learning} Instead of directly differentiating the variational lower bound, one can instead try to optimize the variational posterior directly through dynamic programming.
\begin{align*}
    q^*(\ra_t | \rs_t;\theta) &\approx p(\ra_t | \rs_t, \optimal_{t:T}) \\
    &\approx \frac{p(\optimal_{t:T} | \rs_t,\ra_t)}{p(\optimal_{t:T} | \rs_t)} \\
    &\approx \mathrm{exp}\big( \ln p(\optimal_{t:T} | \rs_t,\ra_t) - \ln p(\optimal_{t:T} | \rs_t) \big)
\end{align*}
Now, using the fact that: 
\begin{align*}
    \ln p(\optimal_{t:T} | \rs_t,\ra_t) &=  \ln \Big( \int d\rs_{t+1} \int d\ra_{t+1} p(\optimal_t | \rs_t)p(\rs_{t+1} |\rs_t,a_t) q(\ra_{t+1} | \rs_{t+1};\theta)p(\optimal_{t+1:T} | \rs_{t+1},\ra_{t+1}) \Big)
\end{align*}
\begin{align*}
    = r(\rs_t,\ra_t) + \ln \E_{q(\ra_{t+1}| \rs_{t+1})p(\rs_{t+1} | \rs_t, \ra_t)}[p(\optimal_{t+1:T} | \rs_{t+1},\ra_{t+1})]
\end{align*}

One can solve this problem recursively by passing backwards messages of the form $p(\optimal_{t:T} | s_t,a_t)$. Intuitively these messages correspond to the probability of acting optimally from the current state and action to the time horizon. These share a close mathematical and intuitive relationship with value functions and state-action value functions, or Q-functions. We define $\mathcal{Q}(\ra_t,\rs_t) = \ln p(\optimal_{t:T} | \rs_t, \ra_t)$ and $\mathcal{V}(\rs_t) = \ln p(\optimal_{t:T} | \rs_t) = \ln \int d\ra_t q(\ra_t | \rs_t)p(\optimal_{t:T} | \rs_t, \ra_t)$. Armed with these definitions, we can write:
\begin{align*}
    q^*(\ra_t | \rs_t) = \exp(\mathcal{Q}(\ra_t,\rs_t) - \mathcal{V}(\rs_t))
\end{align*}
The $\mathcal{Q}$ and $\mathcal{V}$ functions can be computed recursively, which corresponds an iterative message passing algorithm. In a tree-structured MDP, this algorithm corresponds exactly to belief propagation \citep{yedidia2005constructing}. Alternatively, we can \textit{amortise} the computation over a dataset by learning a function $\mathcal{Q}_\phi(\ra_t | \rs_t)$ which maps a state-action pair to a Q-value directly. This function can then be trained on a dataset by a bootstrapping gradient descent.
\begin{align*}
    \frac{d\mathcal{Q}_\phi}{dt} = - \mathbf{E}_{p(\mathcal{D})}\big[ \frac{d\mathcal{Q}_\phi}{d\phi}\Big( \mathcal{Q}_\phi(\rs_t,\ra_t) - r(\rs_t,\ra_t) - \ln E_q[\mathcal{V}(\rs_{t+1})]\Big)\big]
\end{align*}
This update rule differs from the standard Q-learning update in two ways. First, the $\mathcal{Q}$ and $\mathcal{V}$ functions contain action-entropy terms. Secondly, we have a 'soft-max' $\ln \int \cdot$ instead of a 'hard-max' over the next-state value function. As $\beta \rightarrow 0$, the effect of the entropy term and the soft-max will disappear and the standard Q-learning algorithm will be obtained. Moreover, CAI policy-gradients and Q-learning can be combined to yield the soft-actor-critic (SAC) \citep{haarnoja2018soft,haarnoja2018applications} which is a simple, robust, and state-of-the-art model-free algorithm.

\paragraph{Trajectory Optimisation} 
We can also consider directly inferring a \emph{sequence} of actions $q(\ra_{t:T} | \rs_{t:T})$. 
This can be achieved by maximising the variational objective in Eq.\ref{eq:simple-objective} using mirror descent \cite{bubeck2014convex, okada2018acceleration, okada2019variational}, leading to the following iterative update rule:
\begin{equation}
\begin{aligned}
\label{eq:vimpc}
q^{(i+1)}(\rah;\theta) \leftarrow \frac{ q^{(i)}(\rah;\theta) \cdot \mathcal{W}(\rah)\cdot q^{(i)}(\rat;\theta)}{\mathbb{E}_{q^{(i)}(\rat;\theta)}\Big[\mathcal{W}(\rah)\big]\cdot q^{(i)}(\rah;\theta)\Big]} 
\end{aligned}
\end{equation}
where $i$ denotes the current iteration and $\mathcal{W}\big(\rah\big) = \mathbb{E}_{q(\rsh|\rah, \rs_t)}\big[p(\mathcal{O}_{t:T} = 1|\tau)\big]$. 
To infer $q(\rah;\theta)$, Eq. \ref{eq:vimpc} is applied for each state, making this an iterative inference algorithm.
Recent work has demonstrated that Eq. \ref{eq:vimpc} generalises a number of stochastic optimisation methods used extensively in model-based planning \cite{okada2019variational}, including the cross-entropy method (CEM) \citep{rubinstein2013cross} and model-predictive path-integral control (MPPI) \citep{williams2017model}.
Given this generalisation, these methods differ only in their definition of the optimality likelihood $p(\mathcal{O}_{t:T} = 1|\tau)$. 
For instance, CEM defines this quantity as $\mathbf{1}[r(\tau) > r_{\texttt{thd}}]$, where $r_{\texttt{thd}}$ is an arbitrary threshold and $\mathbf{1}$ is the indicator function, while MPPI defines the optimality likelihood as  $\propto \exp\big(r(\tau)\big)$.
Control as inference thus provides a unified perspective on previously unrelated algorithms. 

An alternative approach comes in the form of sequential Monte Carlo (SMC) methods, which provide an elegant approach to probabilistic planning. 
Here, we attempt to approximate the true posterior $p(\rah|\rsh, \mathcal{O}_{t:T}=1)$ with a set of particles $\{\lambda_{t:T}\}$ with weights $\{w_{t:T}\}$. We can derive the update laws for these particles as (see \citet{piche2018probabilistic} for a full derivation). Since the posterior is represented as a set of particles instead of a parametrised distribution, this approach is easily able to handle multimodal posteriors
\begin{align*}
w_t &= w_{t-1} \cdot \frac{p(\rs_t, \ra_t|\rs_{<t}, \ra_{<t}, \mathcal{O}_{t:T})}{q_(\rs_t, \ra_t|\rs_{<t}, \ra_{<t};\theta)} \\ &\propto \ w_{t-1} \cdot \E_{p_{\lambda}(\rs_{t+1}|\rs_t, \ra_t)}\Big[\exp\big(A(\rs_t, \ra_t, \rs_{t+1}\big)\Big]
\end{align*}
Where
\begin{align*}
A(\rs_t, \ra_t, \rs_{t+1}\big) = r(\rs_t, \ra_t) - \log q(\ra_t|\rs_t;\theta)  + \mathcal{V}(\rs_{t+1}) - \log \E_{p(\rs_{t+1}|\rs_t, \ra_t)}\Big[\exp\big(\mathcal{V}(\rs_t)\big)\Big]
\end{align*}
where $\mathcal{V}(\rs_t)$ is a learned value function. Our scheme views this as an iterative planning algorithm, and naturally suggests the idea of amortising the importance weights while also providing the variational lower-bound over the dataset as the principled loss function to optimize against.
\section{Identifying Novel Algorithms}
\label{sec:niovel}

By classifying RL algorithms in terms of amortised vs iterative inference and policy-based vs planning-based, it becomes evident that regions of the design space corresponding to \emph{iterative polices} and \emph{amortised plans} remain relatively unexplored. Next, we discuss the properties and potential implementation of these novel algorithm classes.

\paragraph{Iterative Policies} 
The majority of model-free policy-based algorithms utilize amortised inference. 
However, it is possible to construct a policy-based algorithm that utilises iterative inference, i.e. one that optimises a specific policy or Q-function for each state. This could be achieved by applying policy gradients to simulated trajectories. While this is likely to be inefficient, the two methods could be combined, by initializing the iterative policy with the amortised policy, so that the iterative policy effectively fine-tunes the amortised with respect to the current state and could be used adaptively when the amortised estimate is known to be poor. Interestingly, since the iterative Q-values and policies would be specific to a single state and used for MPC, they do not need to be globally accurate, allowing for more severe approximations than possible with amortised models. iLQR methods \citep{li2004iterative} sit within this quadrant. These methods iteratively infer a policy for each specific state by making a linear approximation to the dynamics and a quadratic approximation to the cost. It would be interesting to combine this intuition with RL by constructing locally approximate Q-values or policies.

\paragraph{Amortised Plans} Planning algorithms generally utilise iterative inference for optimisation, such as by gradient or mirror descent on a variational bound \citep{okada2020planet,srinivas2018universal}. However, one could also construct amortised plans, which learn a global function mapping from states to sequences of actions. Policy gradients present a potential method for learning amortised plans, whereby an approximate posterior over action sequences $q_{\phi}(\rah|\rs_t)$ is optimised using Eq. \ref{eq:policy-gradients}.  
Amortising planning may substantially improve the computational efficiency of MPC algorithms, especially when adaptively combined with iterative planning, so that expensive online iterative planning is only used where absolutely necessary. Amortised plans correspond to the notion of fixed-action-patterns in the study of biological behaviour \citep{lorenz2013foundations}, and could enable agents to learn temporally extended 'macro-actions' .

\section{Conclusion}
\label{sec:conclusion}

We have explored a novel classification scheme for RL algorithms, based on iterative and amortised variational inference within the mathematically principled control-as-inference framework. Our scheme informatively partitions a range of influential RL approaches, including policy gradients, Q-learning, actor-critic methods, and trajectory optimisation algorithms such as CEM, MPPI and SMC -- highlighting relationships which would have otherwise remained obscure. We have shown how constructing algorithmic design-spaces based on fundamental distinctions can reveal unexplored design choices. Our work highlights the importance of identifying the common factors underlying disparate RL algorithms and the utility of building unifying conceptual frameworks through which to understand them. 

Future work may explore still other classification schemes that can be derived from the perspective of control as inference.
For instance, algorithms can be classified in terms of whether the approximate posterior $q(\ra_t|\rs_t)$ is parametric or non-parametric \citep{marinodesign}, and whether the action prior $p(\ra_t)$ is learned or uniform \citep{marinoinference} , and whether variational, dynamic-programming, or importance-sampling inference methods are used. Ultimately, by fully quantifying and classifying the relevant axes of variation, we hope to develop a unified understanding of the design-space of RL algorithms, which would be instrumental in situating, clarifying, and inspiring future research.

Finally, our work illuminates the possibility of \textit{combining} iterative and amortised inference \citep{marino2018iterative}. This approach has been explored in the context of unsupervised learning, where a hybrid approach to inference can help overcome the shortcomings of using either iterative or amortised inference alone (Tschantz et al 2020, in press). Given the correspondence between iterative \& amortised inference and planning \& policies, our scheme suggests a potential avenue towards combining the sample efficiency of model-based planning and the asymptotic performance of model-free policy optimisation in a mathematically principled manner.

\nocite{baltieri2019pid,watkins1992q,theodorou2010reinforcement,hansen2003reducing,williams1992simple,auger2005restart,rubinstein1997optimization}
\bibliography{refs}
\newpage

\begin{figure}
    \begin{center}
          \includegraphics[width=0.35\textwidth]{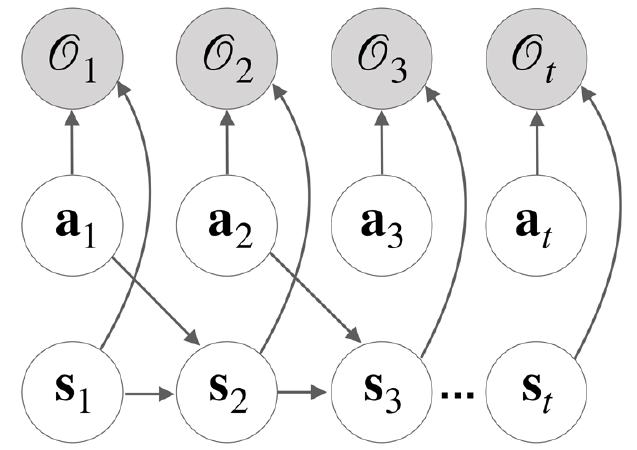}
    \end{center}
    \caption{Graphical model for control as inference, with optimality variables $\optimal$.}
\label{fig:graphical-model}
\end{figure}

\section*{Appendix A: Bound derivations}
\label{ap:bound}

We wish to minimize the KL divergence between approximate and true posterior. Since this divergence is intractable (it contains the true posterior which is intractable), we instead show that the divergence between the approximate posterior and generative model (which is tractable) lower-bounds the divergence we want. Thus, by maximizing this lower-bound, we bring the true and approximate posteriors closer together.

\begin{align*}
    &\KL \Big(q(\tau;\theta) \Vert p(\tau | \optimal_{t:T}) \Big) = \mathbb{E}_{q(\tau ; \theta)} \big[ \ln q(\tau; \theta) - \ln p(\tau | \optimal_{t:T}) \big] \\
    &= \mathbb{E}_{q(\tau ; \theta)} \big[ \ln q(\tau; \theta) - \ln p(\tau | \optimal_{t:T}) + \ln p(\optimal_{t:T}) - \ln p(\optimal_{t:T}) \big] \\
    &= \mathbb{E}_{q(\tau ; \theta)} \big[ \ln q(\tau; \theta) - \ln p(\tau, \optimal_{t:T}) + \ln p(\optimal_{t:T}) \big] \\
    &= \mathbb{E}_{q(\tau ; \theta)} \big[ \ln q(\tau; \theta) - \ln p(\tau, \optimal_{t:T}) \big] + \ln p(\optimal_{t:T}) \\
    &\implies \KL \Big(q(\tau;\theta) \Vert p(\tau | \optimal_{t:T}) \Big) \geq \underbrace{\KL \Big(q(\tau;\theta) \Vert p(\tau, \optimal_{t:T}) \Big)}_{\text{$\mathcal{L}$}}
\end{align*}

Now, if we substitute in the definitions of the approximate posterior and generative model from the main text, we obtain.
\begin{align*}
\mathcal{L} &= \KL \Big(q(\tau;\theta) \Vert p(\tau, \optimal_{t:T}) \Big) = \mathbb{E}_{q(\tau ;\theta)}\big[  \ln \big( q(\rs_1) \prod^T_{t=1} q(\rs_{t+1}|\rs_t, \ra_t)q(\ra_t|\rs_t;\theta) \big) -\ln \big( p(\rs_1) \prod^T_{t=1} p(\mathcal{O}_t|\rs_t, \ra_t) p(\rs_{t+1}|\rs_t, \ra_t)p(\ra_t) \big) \big] \\
&= \sum_t^T \, \underbrace{\mathbb{E}_{q(\ra_t, \rs_t; \theta)} \big[ p(\optimal_t | \rs_t, \ra_t) \big]}_{\text{Expected Reward}}  - \underbrace{\mathbb{E}_{q(\ra_{t-1} | \rs_{t-1};\theta)} \big[ KL \Big(q(\rs_t | \rs_{t-1}, \ra_{t-1}) \Vert p(\rs_t | \rs_{t-1}, \ra_{t-1}) \Big)\big]}_{\text{State Complexity}} 
\\ &- \underbrace{\mathbb{E}_{q(\rs_t | \rs_{t-1} \ra_{t-1})}\big[ \KL \Big(q(\ra_t | \rs_t; \theta) \Vert p(\ra_t | \rs_t) \Big) \big]}_{\text{Action Complexity}}
\end{align*}

We see that the CAI objective breaks down into three separate terms. The first, the expected reward, quantifies the expected sum of rewards an agent is likely to obtain for a given trajectory. The state-complexity term penalizes trajectories where the approximate and prior trajectories differ, while the action-trajectory term penalizes the divergence between the agent's actions and some prior action distribution. If we assume, as is commonly done in the literature that the approximate and generative dynamics are the same: $q(\rs_t | \rs_{t-1}, \ra_{t-1}) \eqdef p(\rs_t | \rs_{t-1},\ra_{t-1})$, then the state-complexity term vanishes. This is a well motivated assumption, since having separate approximate dynamics effectively means the agent thinks it has control over the environmental dynamics, and will thus tend towards risk-seeking policies if it does not actually have this degree of control over its environment. 

The CAI framework also often ignores the action prior $p(\ra_t | \rs_t)$. This does not necessarily lead to a lack of generality since the action prior can always be subsumed into the reward. Nevertheless, it is often intuitively useful to think of utilising the action prior in some way. For instance, in many control tasks, action itself is costly. For instance, consider the task of flying a rocket. Actions such as applying thrust deplete fuel, and thus have a cost associated with them which can be well-modelled with an action prior of 0 (any action at all incurs a small penalty).

The action prior also provides a mathematically principled way to combine iterative and amortised inference. Suppose that we optimize the iterative bound $\mathcal{L}(\theta)$ for each datapoint, but we also have a trained amortised policy $q_\phi(\ra_t | \rs_t)$, then we can set the action prior to be the output of the amortised scheme $p(\ra_t | \rs_t) \eqdef q_\phi(\ra_t | \rs_t)$ and infer the iterative posterior $q(\ra_t | \rs_t; \theta)$ using this prior.

If, as is commonly done we ignore the action prior by assuming it is uniform -- $p(\ra_t | \rs_t) \eqdef \frac{1}{|\mathcal{A}|}$ in discrete state-spaces and $p(\ra_t | \rs_t) \eqdef \text{\textit{Unif}}(\ra_{min},\ra_{max})$ in continuous action spaces, then the action-complexity term disappears and we obtain for the bound.
\begin{align*}
    \mathcal{L} &= \sum_t^t \, \mathbb{E}_{q(\ra_t | \rs_t)p(\rs_t | \rs_{t-1},\ra_{t-1})}\big[ \ln p(\optimal_{t} | \rs_t \ra_t) - \ln q(\ra_t | \rs_t) \big] \\
    &= \sum_t^t \, \mathbb{E}_{q(\ra_t | \rs_t)p(\rs_t | \rs_{t-1},\ra_{t-1})}\big[ \ln p(\optimal_{t} | \rs_t \ra_t) \big] + \mathcal{H}\Big[q(\ra_t |\rs_t)\Big]
\end{align*}

Which corresponds exactly to the bound given in Equation 2 of the main text.

\end{document}